\documentclass[lettersize,journal]{IEEEtran}
\usepackage{amsmath,amsfonts}
\usepackage{algorithmic}
\usepackage{array}
\usepackage[caption=false,font=normalsize,labelfont=sf,textfont=sf]{subfig}
\usepackage{textcomp}
\usepackage{stfloats}
\usepackage{url}
\usepackage{verbatim}
\usepackage{graphicx}
\def\BibTeX{{\rm B\kern-.05em{\sc i\kern-.025em b}\kern-.08em
    T\kern-.1667em\lower.7ex\hbox{E}\kern-.125emX}}
\usepackage{balance}
\usepackage{subfiles}
\usepackage{booktabs}
\usepackage{multirow}
\usepackage{cite}
\usepackage[table]{xcolor}
\usepackage{makecell}
\usepackage{bm}
\usepackage[switch]{lineno}
\usepackage{amssymb}

\usepackage{makecell}
\usepackage{multirow}

\definecolor{nbarrier}{RGB}{255, 120, 50}
\definecolor{nbicycle}{RGB}{255, 192, 203}
\definecolor{nbus}{RGB}{255, 255, 0}
\definecolor{ncar}{RGB}{0, 150, 245}
\definecolor{nconstruct}{RGB}{0, 255, 255}
\definecolor{nmotor}{RGB}{200, 180, 0}
\definecolor{npedestrian}{RGB}{255, 0, 0}
\definecolor{ntraffic}{RGB}{255, 240, 150}
\definecolor{ntrailer}{RGB}{135, 60, 0}
\definecolor{ntruck}{RGB}{160, 32, 240}
\definecolor{ndriveable}{RGB}{255, 0, 255}
\definecolor{nother}{RGB}{139, 137, 137}
\definecolor{nsidewalk}{RGB}{75, 0, 75}
\definecolor{nterrain}{RGB}{150, 240, 80}
\definecolor{nmanmade}{RGB}{213, 213, 213}
\definecolor{nvegetation}{RGB}{0, 175, 0}

\definecolor{nvcolor}{RGB}{119,185,0}
\definecolor{roadcolor}{RGB}{234,51,246}
\definecolor{sidewalkcolor}{RGB}{68,8,72}
\definecolor{parkingcolor}{RGB}{241,156,249}
\definecolor{othergroundcolor}{RGB}{160,32,76}
\definecolor{buildingcolor}{RGB}{246,202,69}
\definecolor{carcolor}{RGB}{111,149,238}
\definecolor{truckcolor}{RGB}{74,32,172}
\definecolor{bicyclecolor}{RGB}{136,227,242}
\definecolor{motorcyclecolor}{RGB}{37,59,146}
\definecolor{othervehiclecolor}{RGB}{96,81,242}
\definecolor{vegetationcolor}{RGB}{79, 173, 50}
\definecolor{trunkcolor}{RGB}{126, 65, 22}
\definecolor{terraincolor}{RGB}{171, 238, 105}
\definecolor{personcolor}{RGB}{234, 60, 49}
\definecolor{bicyclistcolor}{RGB}{234, 66, 195}
\definecolor{motorcyclistcolor}{RGB}{138, 42, 90}
\definecolor{fencecolor}{RGB}{238, 128, 69}
\definecolor{polecolor}{RGB}{252, 241, 161}
\definecolor{trafficsigncolor}{RGB}{233, 51, 35}
\definecolor{other-struct.color}{RGB}{255, 150, 0}
\definecolor{other-objectcolor}{RGB}{50, 255, 255}
\definecolor{lane-markingcolor}{RGB}{150, 255, 170}
\definecolor{color1}{RGB}{176, 36, 24}
\definecolor{color2}{RGB}{0, 176, 80}
\definecolor{color3}{RGB}{0, 0, 200}

\DeclareSubrefFormat{parens}{#1(#2)}

\begin{document}

\title{SuperOcc: Toward Cohesive Temporal Modeling for Superquadric-based 3D Occupancy Prediction}
\author{Zichen~Yu,
        Quanli~Liu,~\IEEEmembership{Member,~IEEE},
        Wei~Wang,~\IEEEmembership{Senior Member,~IEEE}, \\
        Liyong~Zhang,~\IEEEmembership{Member,~IEEE},
        and Xiaoguang~Zhao
        \IEEEcompsocitemizethanks{
            \IEEEcompsocthanksitem~Zichen Yu, Quanli Liu, Wei Wang and Liyong Zhang are with the School of Control Science and Engineering, 
            Dalian University of Technology, Dalian 116024, China, and also with the Dalian Rail Transmit Intelligent Control and 
            Intelligent Operation Technology Innovation Center, Dalian 116024, China (e-mail: yuzichen@mail.dlut.edu.cn; liuql@dlut.edu.cn; 
            wangwei@dlut.edu.cn; zhly@dlut.edu.cn).
            \IEEEcompsocthanksitem~Xiaoguang Zhao is with the Dalian Rail Transmit Intelligent Control and Intelligent Operation Technology 
            Innovation Center, Dalian 116024, China, and also with the Dalian Seasky Automation Co., Ltd, Dalian 116024, China (e-mail: xiaoguang.zhao@dlssa.com).
            \IEEEcompsocthanksitem~This work was supported in part by the National Natural Science Foundation of China under Grant 62373077, 
            in part by the Science and Technology Joint Plan of Liaoning Province under Grant 2024JH2/102600014, in part by the Fundamental Research 
            Funds for the Central Universities under Grant DUTZD25203 and in part by the Key Field Innovation Team Support Plan of Dalian, China, 
            under Grant 2021RT02.
            \textit{(Corresponding author: Quanli Liu).} 
        }
}

\maketitle

\begin{abstract}
3D occupancy prediction plays a pivotal role in the realm of autonomous driving, as it provides a 
comprehensive understanding of the driving environment.  Most existing methods construct dense scene 
representations for occupancy prediction, overlooking the inherent sparsity of real-world driving 
scenes. Recently, 3D superquadric representation has emerged as a promising sparse alternative to 
dense scene representations due to the strong geometric expressiveness of superquadrics. However, 
existing superquadric frameworks still suffer from insufficient temporal modeling, a challenging 
trade-off between query sparsity and geometric expressiveness, and inefficient superquadric-to-voxel 
splatting. 
To address these issues, we propose SuperOcc, a novel framework for superquadric-based 3D occupancy 
prediction. SuperOcc incorporates three key designs: (1) a cohesive temporal modeling mechanism to 
simultaneously exploit view-centric and object-centric temporal cues; (2) a multi-superquadric 
decoding strategy to enhance geometric expressiveness without sacrificing query sparsity; and (3) an 
efficient superquadric-to-voxel splatting scheme to improve computational efficiency. Extensive 
experiments on the SurroundOcc and Occ3D benchmarks demonstrate that SuperOcc achieves 
state-of-the-art performance while maintaining superior efficiency.
The code is available at https://github.com/Yzichen/SuperOcc.
\end{abstract}

\section{Introduction}
\label{sec:intro}

\IEEEPARstart{V}{ision-centric} 3D scene understanding is a core research direction in autonomous driving. Existing approaches primarily 
focused on 3D object detection~\cite{huang2021bevdet, li2022bevformer, liu2022petr,liu2023petrv2, liu2023sparsebev}, predicting 3D bounding 
boxes of traffic participants to provide object-level representations. 
However, they struggle to handle out-of-vocabulary or arbitrarily shaped objects, which significantly compromises the safety and 
reliability of driving systems~\cite{tian2023occ3d, tong2023scene}. To overcome this limitation, 3D occupancy prediction divides the scene into voxel grids and 
jointly predicts the occupancy state and semantic label of each voxel. Compared with 3D object detection, occupancy prediction 
offer a unified representation of all scene elements, enabling more comprehensive and fine-grained scene understanding.

\begin{figure}[t]%
  \centering
  \captionsetup[subfloat]{font=footnotesize, labelfont=footnotesize}
  \subfloat[{RayIoU vs. FPS on Occ3D Benchmark}]{
    \includegraphics[width=0.8\linewidth]{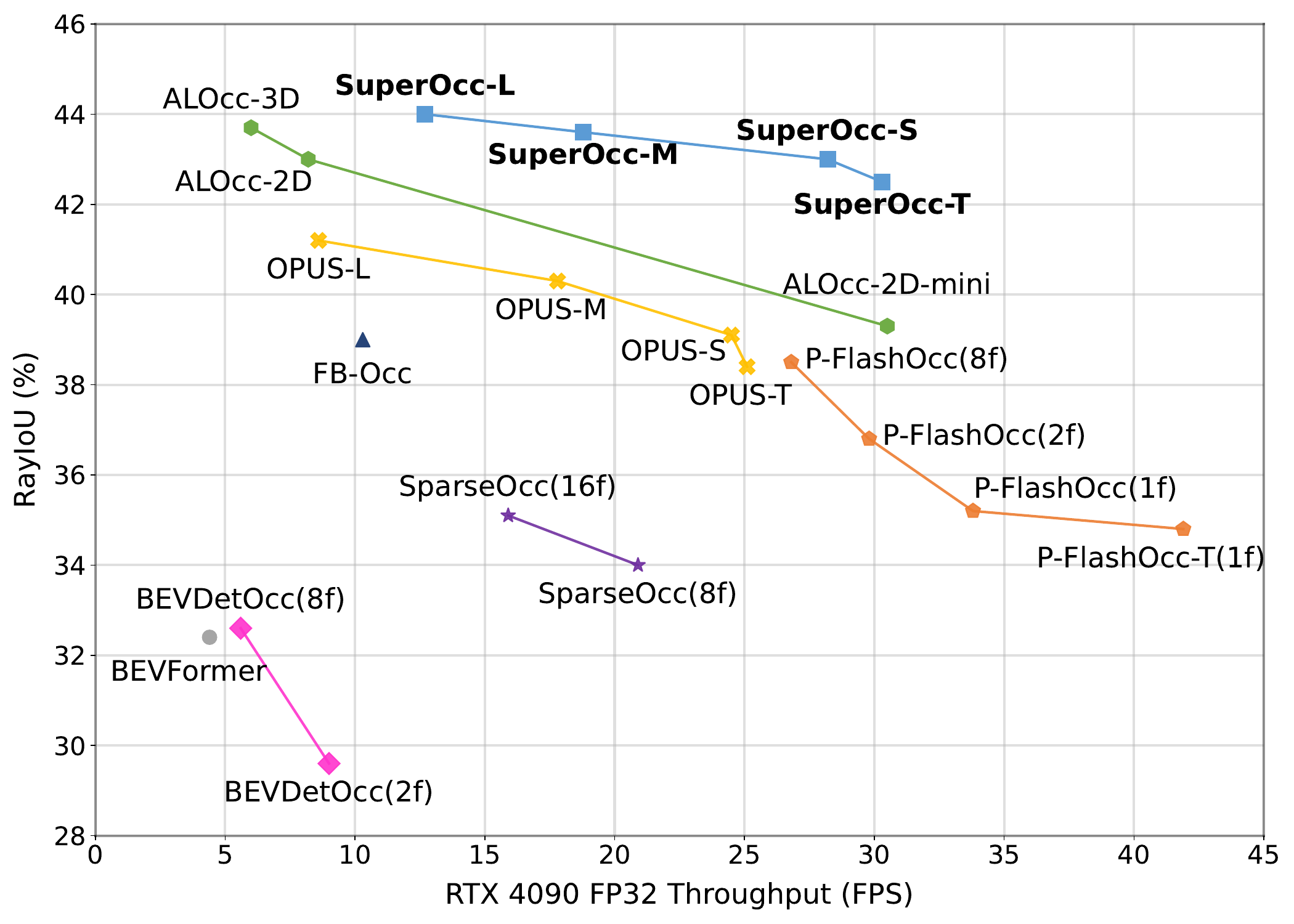}
    \label{fig:occ3d}
    }\hfill
  \subfloat[{mIoU vs. FPS on SurroundOcc Benchmark}]{
    \includegraphics[width=0.8\linewidth]{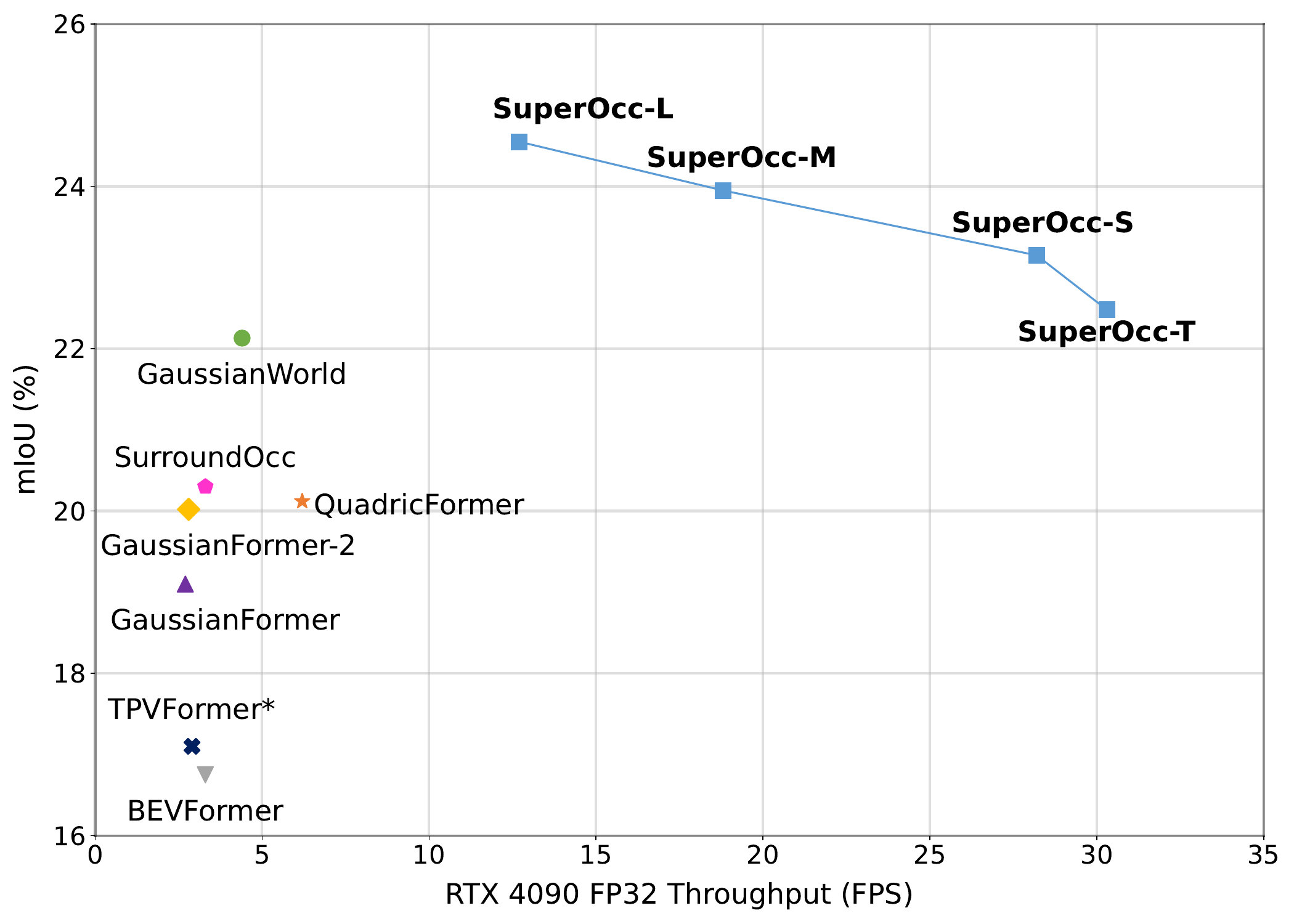}
    \label{fig:surroundocc}
    }\\    
  \caption{
    Comparison of speed-accuracy trade-offs of different methods on (a) Occ3D and (b) SurroundOcc benchmarks.
    SuperOcc achieves the optimal trade-off, delivering high accuracy while maintaining efficient inference.
  }
  \label{fig:trade_off}
\end{figure}

3D occupancy prediction is essentially a 3D semantic segmentation task. Therefore, early methods~\cite{li2023fbocc, wei2023surroundocc, zhang2023occformer, tong2023scene, tian2023occ3d} 
directly generate dense volumetric feature representations followed by per-voxel classification. These methods are intuitive and 
capable of capturing sufficient geometric details, but they suffer from high computational complexity and memory consumption. 
To alleviate this burden, subsequent studies have extensively explored alternative dense representations~\cite{yu2023flashocc, yu2024panoptic, huang2023tri} 
and more efficient learning schemes~\cite{wang2024panoocc, ma2024cotr}. 
Although these methods effectively reduce computational costs, they still fail to fully exploit the spatial sparsity of real-world scenes, i.e., the 
vast majority of 3D space is actually free. Consequently, performing uniform computation and prediction across these free regions introduces considerable 
computational redundancy and resource waste. 

To align with the inherent sparsity of driving scenes, recent research~\cite{liu2024sparseocc, wang2024opus, huang2024gaussianformer, huang2025gaussianformer, zuo2025gaussianworld, zuo2025quadricformer} 
has begun exploring sparse scene representations. 
Gaussian-based methods~\cite{huang2024gaussianformer, huang2025gaussianformer} describe scenes sparsely using a set of 3D semantic 
gaussians, and each of them is responsible for modeling a flexible region of interest. However, the strong elliptical priors limit 
their ability to model diverse geometric shapes. 
Superquadric-based methods~\cite{zuo2025quadricformer, hayes2025superquadricocc} address this limitation by introducing geometrically 
expressive superquadrics as scene primitives. With a compact parametric representation, superquadrics can generate a rich variety 
of geometric shapes, allowing complex structures to be modeled effectively without dense stacking. 
Despite the significant progress made by superquadric-based methods in scene representation, they still suffer from the following key 
limitations:

\begin{itemize}
\item \textbf{Insufficient Temporal Modeling}.
Existing temporal-aware methods are restricted to a singular modeling paradigm, i.e., view-centric or object-centric temporal modeling. 
The former exploits historical image features to provide fine-grained spatio-temporal context, while the latter propagates temporal information
via sparse queries focusing on high-level semantic and geometric abstractions. The absence of a unified temporal modeling strategy limits 
the effective utilization of temporal cues in dynamic scenes.

\item \textbf{Challenging Trade-off between Query Sparsity and Geometric fidelity}.
Current frameworks decode each query into a single superquadric primitive, which severely constrains the geometric expressiveness of 
individual queries. As a result, it is difficult to accurately reconstruct complex driving scenarios with a limited number of queries, 
forcing a challenging trade-off between geometric fidelity and query sparsity.

\item \textbf{Inefficient Superquadric-to-Voxel Splatting}. 
The implementation of the superquadric-to-voxel splatting lacks operator-level optimization tailored for the CUDA architecture, resulting 
in redundant memory usage and inefficient computation.
This constrains the model's deployment potential in resource-constrained in-vehicle systems.
\end{itemize}

To address these limitations, we propose a novel framework SuperOcc for superquadric-based 3D occupancy prediction. 
It mainly contains three key designs:
1)~The \textbf{cohesive temporal modeling mechanism} is developed to simultaneously leverage view-centric and object-centric temporal modeling within a unified framework. 
For the former, we maintain a memory queue to cache historical image features, which are sampled and aggregated to provide fine-grained spatio-temporal context. 
For the latter, sparse queries serve as hidden states for temporal propagation. Historical queries are stored in a memory queue and propagated to the subsequent 
frame following temporal alignment. They can provide informative spatial and semantic priors for subsequent predictions, since the evolution of the driving 
scenarios is continuous. This dual-path integration enables the model to comprehensively exploit temporal cues, improving both the accuracy and stability of 
occupancy predictions in dynamic driving scenarios.
2)~The \textbf{multi-superquadric decoding strategy} is designed to enhance the model's geometric expressiveness while preserving query sparsity. 
Built upon the superquadric-based scene representation, this strategy decodes each query into a local cluster of superquadrics in a coarse-to-fine manner. 
This allows a single query to model intricate geometric structures with high fidelity. 
Consequently, complex 3D scenes can be accurately represented even with only a few hundred sparse queries.
3)~The \textbf{efficient superquadric-to-voxel splatting scheme} is implemented to accelerate the transformation from 3D superquadric representations to 3D semantic occupancy. 
By introducing a tile-level binning strategy and leveraging GPU shared memory, this optimization substantially reduces computational complexity and memory access overhead, 
significantly improving both training and inference efficiency.

Overall, our contributions are summarized as follows:
\begin{itemize}
\item We propose SuperOcc, a novel 3D occupancy prediction framework that introduces a cohesive temporal modeling mechanism. 
This mechanism captures fine-grained spatio-temporal context from historical image features and efficiently leverages informative historical 
priors through query propagation, thereby enhancing occupancy modeling accuracy in dynamic driving scenarios.


\item We design a multi-superquadric decoding strategy to boost the geometric expressiveness of sparse queries, alongside an efficient 
superquadric-to-voxel splatting implementation to significantly reduce training costs and improve inference efficiency.

\item We evaluate SuperOcc on the challenging Occ3D~\cite{tian2023occ3d} and SurroundOcc~\cite{wei2023surroundocc} benchmarks, achieving state-of-the-art~(SOTA)
performance while maintaining superior efficiency, as illustrated in Fig.~\ref{fig:trade_off}.

\end{itemize}

\section{Related Works}
\label{sec:related-work}

\subsection{Camera-based 3D Occupancy Prediction}
\subsubsection{Dense Scene Representation}
Dense scene representation is the most straightforward modeling paradigm. Existing methods~\cite{cao2022monoscene, li2023fbocc, wei2023surroundocc, zhang2023occformer, 
tian2023occ3d, ma2024cotr, wang2024panoocc, li2023voxformer, oh20253d, kim2025protoocc, tan2025GEOcc, Ren2025RM2Occ} typically discretizes the 3D space into voxel grids to construct a dense volume representation.
FB-OCC~\cite{li2023fbocc} employs depth-aware back-projection to complete and enhance the initial 3D volume generated by forward projection. 
SurroundOcc~\cite{wei2023surroundocc} adopts a 2D-3D spatial attention mechanism to integrate multi-camera information into 3D volume queries and 
construct 3D volume features in a multi-scale manner. Although these methods are capable of capturing rich geometric details, 
they suffer from high computational cost and memory overhead. Subsequent works therefore aim to improve the efficiency of 
dense volume construction and processing. OccFormer\cite{zhang2023occformer} decomposes heavy 3D processing into local and global transformer 
pathways along the horizontal plane to efficiently process the 3D volume. CTF-Occ~\cite{tian2023occ3d} introduces an incremental token selection 
strategy to mitigate the computational burden caused by dense sampling. To better balance performance and efficiency, COTR~\cite{ma2024cotr}, 
PanoOcc~\cite{wang2024panoocc} first generate a compact low-resolution volume and then recover the geometric details via voxel upsampling.

In addition to directly constructing the 3D volume, some methods~\cite{huang2023tri, yu2023flashocc, yu2024panoptic, zhang2024lightweight, hou2024fastocc} seek more efficient dense representations. TPVFormer~\cite{huang2023tri} proposes
a Tri-perspective view~(TPV) representation, where each TPV plane aggregates image features through deformable attention, and 
each voxel is modeled as the sum of its projected features onto the three planes. To further enhance deployment friendliness, 
FlashOcc~\cite{yu2023flashocc} compresses the 3D scene into a BEV representation and adopts a channel-to-height decoding strategy to directly infer 
voxel-level occupancy from it.

\subsubsection{Sparse Scene Representation}
Given the spatial sparsity of driving scenarios, sparse scene representation~\cite{liu2024sparseocc, wang2024opus, huang2024gaussianformer, huang2025gaussianformer, 
zuo2025quadricformer, zuo2025gaussianworld, liao2025stcocc, shi2024occupancy} focuses on modeling non-free areas, thereby avoiding redundant computations for large free spaces. 
SparseOcc~\cite{liu2024sparseocc} proposes a sparse voxel decoder that progressively reconstructs 
sparse 3D voxel representations of occupied regions through multi-stage sparsification and refinement. However, the 
sparsification process heavily relies on the accurate estimation of voxel occupancy scores. OPUS~\cite{wang2024opus} predicts the locations and 
semantic classes of occupied points with a set of learnable queries, eliminating the need for explicit space modeling or complex 
sparsification procedures. Recent advances adopt parametric geometric primitives as the fundamental units for scene 
representation. GaussianFormer~\cite{huang2024gaussianformer} and GaussianFormer-2~\cite{huang2025gaussianformer} utilize a set of 3D semantic gaussians to sparsely describe a 3D scene, 
where each gaussian is responsible for modeling a flexible region of interest. However, the inherent ellipsoidal shape priors limit their ability to model diverse structures.
Therefore, QuadricFormer~\cite{zuo2025quadricformer} further proposes the 3D semantic superquadric
representation, which can efficiently represent complex structures with fewer primitives due to the stronger geometric 
expressiveness of superquadrics.

\subsection{Camera-based Temporal Modeling}
Camera-based temporal modeling aims to exploit historical observations to improve spatial understanding and temporal consistency in 3D perception. 
Existing methods can be broadly categorized into view-centric and object-centric temporal modeling paradigms.

The view-centric paradigm extracts historical information from structured scene representations across frames, such as perspective view~(PV), 
bird's-eye-view~(BEV), or 3D volume representations. 
Specifically, PV-based methods~\cite{liu2023petrv2, liu2023sparsebev, lin2022Sparse4D, liu2024sparseocc, wang2024opus} typically perform temporal modeling 
by interacting multi-frame PV features with object queries.
For example, PETRv2~\cite{liu2023petrv2} utilizes 3D position embedding to transform multi-frame, multi-view 2D features into 3D position-aware features, and 
then object queries interact with these features using global cross-attention. 
To improve the interaction efficiency, SparseBEV~\cite{liu2023sparsebev}, Sparse4D~\cite{lin2022Sparse4D}, SparseOcc~\cite{liu2024sparseocc}, OPUS~\cite{wang2024opus} adopt 
adaptive sparse sampling and aggregation. In parallel, BEV/volume-based methods~\cite{li2022bevformer, park2022solofusion, wang2024panoocc, he2025gsdocc, chen2025alocc, chen2025rethinking, li2024viewformer, leng2025occupancy, zong2023temporal, xia2024henet, chang2024recurrentbev, Han2024videobev} focus on the 
fusion of multi-frame BEV/volume features. 
SOLOFusion~\cite{park2022solofusion}, PanoOcc~\cite{wang2024panoocc}, GSD-Occ~\cite{he2025gsdocc} and ALOcc~\cite{chen2025alocc} align historical features to the current frame based on ego-motion, and then fuse them with current features via feature concatenation. 
BEVFormer~\cite{li2022bevformer} and ViewFormer~\cite{li2024viewformer} utilize temporal self-attention to extract temporal information from historical BEV features.

The object-centric paradigm propagates temporal information with sparse queries. Representative methods such as StreamPETR~\cite{wang2023exploring}, Sparse4Dv2~\cite{lin2023sparse4dv2}, Sparse4Dv3~\cite{lin2023sparse4dv3} 
propagate temporal information through query propagation and temporal cross-attention between current and historical queries.

However, these two paradigms are typically explored independently. View-centric methods capture fine-grained spatio-temporal context, whereas object-centric methods 
efficiently propagate historical priors. To leverage their complementary strengths, we propose a 
cohesive temporal modeling mechanism that jointly exploits view-centric and object-centric temporal cues 
within a unified framework.

\begin{figure*}[t]
    \centering
    \includegraphics[width=\linewidth]{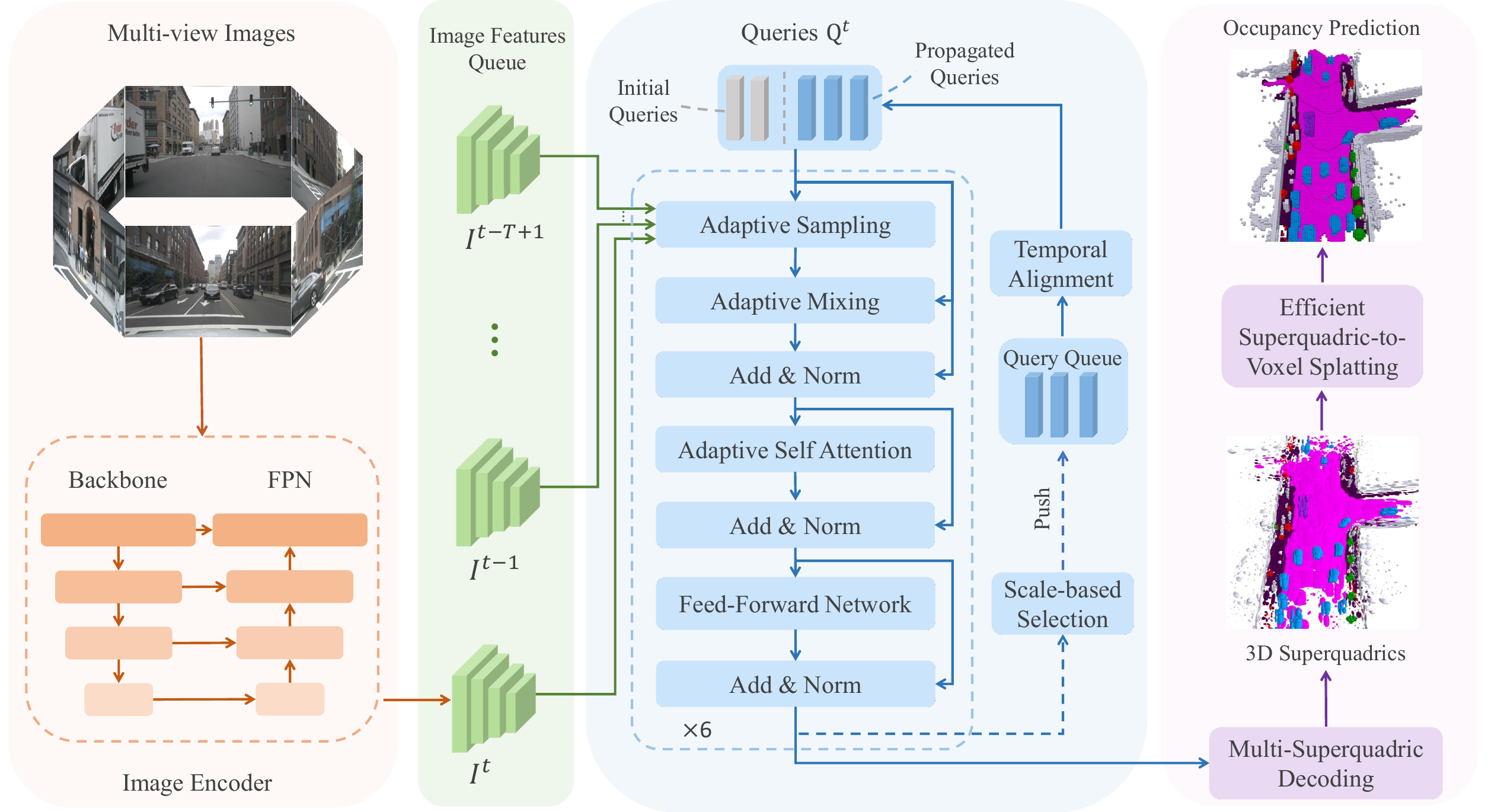}
    \caption{Framework of SuperOcc. Given multi-view image sequences, SuperOcc constructs a superquadric-based sparse scene representation for 
    3D occupancy prediction. To achieve comprehensive temporal modeling, the framework extracts fine-grained spatio-temporal context through 
    the interaction of sparse queries with multi-frame image features, while efficiently exploiting informative historical priors via query propagation. 
    Furthermore, each updated query is decoded into a set of semantic superquadrics through a multi-superquadric decoding strategy. 
    Finally, voxel-level occupancy prediction is generated through an efficient superquadric-to-voxel splatting process.
    }
    \label{fig:framework}
\end{figure*}

\section{Method}

\subsection{Preliminaries: 3D Semantic Superquadric Representation}
The 3D semantic superquadric representation utilizes a set of general posed superquadric primitives to describe a 3D scene: 
\begin{equation}
    \mathcal{S} = \{\mathbf{S}^i\}_{i=1}^M = \left\{ \bigl(\textbf{m}^i, \textbf{r}^i, \textbf{s}^i, \bm{\epsilon}^i, \sigma^i, \textbf{c}^i \bigr) \right\}_{i=1}^M,
\end{equation}
where the center $\mathbf{m}=(m_x, m_y, m_z)^T$ and rotation $\mathbf{r}\in \mathbb{R}^4$ specify the pose of the superellipsoid.  $\mathbf{s}=(s_x,s_y,s_z)^T$ determines 
the scaling factors along the three axes, while the squareness parameter $\bm{\epsilon} = (\epsilon_{1}, \epsilon_{2})^T$ controls the squareness of the shape. 
Opacity $\sigma$ and semantic probability $\mathbf{c}$ are equipped to incorporate semantic information.

QuadricFormer~\cite{zuo2025quadricformer} interprets each superquadric as a continuous occupancy field that describes the occupancy probability in its local neighborhood rather than a hard surface boundary. 
Specifically, for a 3D point $\textbf{p}$, the probability of it being occupied by the superquadric $\mathbf{S}$ is defined as:
\begin{equation}
\scalebox{0.85}{$
p_o(\mathbf{p}; \mathbf{S}) 
= \exp\Biggl(-\lambda \Biggl( 
    \Biggl( \left(\tfrac{|x'|}{s_x}\right)^{\tfrac{2}{\epsilon_2}} 
      + \left(\tfrac{|y'|}{s_y}\right)^{\tfrac{2}{\epsilon_2}} 
    \Biggr)^{\tfrac{\epsilon_2}{\epsilon_1}} 
    + \left(\tfrac{|z'|}{s_z}\right)^{\tfrac{2}{\epsilon_1}} 
\Biggr)\Biggr),
$}
\end{equation}
where $\lambda$ is a temperature parameter controlling the decay rate of the occupancy probability, and $\mathbf{p}'=(x',y',z')^T$ denotes the coordinates 
of point $\mathbf{p}$ in the local coordinate system of the superquadric. The local coordinates are obtained via a rotation and translation:
\begin{equation}
    \mathbf{p}'=\mathbf{R}^T(\mathbf{p}-\mathbf{m}),
\end{equation}
where $\mathbf{R} \in SO(3)$ is the rotation matrix derived from the rotation $\mathbf{r}$. 
Assuming that the occupancy events associated with different superquadrics at point $\mathbf{p}$ are mutually independent, 
the overall occupancy probability is derived as the complement of the joint non-occupancy probability:
\begin{equation}
    p_o(\mathbf{p}) = 1 - \prod_{i=1}^{M} \left( 1 - p_o(\mathbf{p}; \mathbf{S}^i) \right).
\end{equation}
The semantic prediction of the point $\mathbf{p}$ is formulated as a weighted aggregation of the semantic predictions of all contributing superquadrics:
\begin{equation}
    \mathbf{p_s}(\mathbf{p}) = \frac{\sum_{i=1}^M p_o(\mathbf{p}; \mathbf{S}^i)\, \sigma^i\mathbf{c}^i}{\sum_{j=1}^M p_o(\mathbf{p}; \mathbf{S}^j)\sigma^j}.
\end{equation}
Finally, the semantic occupancy prediction is generated as:
\begin{equation}
    \mathbf{o}(\mathbf{p})=[p_o(\mathbf{p}) \cdot \mathbf{p_s}(\mathbf{p}); 1-p_o(\mathbf{p})],
\end{equation}
where $p_o(\mathbf{p})$ weights the semantic prediction and $1-p_o(\mathbf{p})$ corresponds to the probability of the empty class.

\subsection{Overall framework}
As illustrated in Fig.~\ref{fig:framework}, SuperOcc follows an encoder-decoder architecture. The image encoder comprises a backbone~(e.g., ResNet50~\cite{he2016deep}) and a neck~(e.g., FPN~\cite{lin2017feature}). 
Given multi-view images at time $t$ as input, the image encoder extracts multi-scale features denoted as $I^t=\{ I^{t,v,l} \in \mathbb{R}^{C \times H_l \times W_l}|1 \leq l \leq L, 1 \leq v \leq V\}$, 
where $V$, $L$, and $C$ represent the number of camera views, feature scales, and feature channels, respectively.
$H_l$ and $W_l$ denote the feature resolution at the $l$-th feature scale.
To leverage temporal information, the image features from the most recent $T$ frames are maintained in a memory queue, 
denoted as $\mathcal{I}=\{I_{t}, I_{t-1}, \cdots, I_{t-T+1}\}$. 

Subsequently, we initialize a set of queries $Q = \{\mathbf{q}^i \in \mathbb{R}^C \}_{i=1}^{N_q}$ and a corresponding set of 3D reference points 
$P_{ref} = \{\mathbf{p}_{ref}^i \in \mathbb{R}^3 \}_{i=1}^{N_q}$ in the decoder. To achieve comprehensive temporal modeling, SuperOcc integrates 
view-centric and object-centric paradigms within a unified framework. 
Specifically, the view-centric path adaptively samples and aggregates multi-frame image features from a memory queue to capture detailed 
spatio-temporal cues guided by the queries. 
Simultaneously, the object-centric path propagates historical queries to the subsequent frame, enabling the preservation and reuse of object-level
semantic and spatial priors across frames.
These queries then interact globally through adaptive self-attention~\cite{liu2023sparsebev}.
Finally, the updated query features are decoded into geometric parameters and semantic logits for a set of superquadrics, which are then 
transformed into voxel-level occupancy prediction via an efficient superquadric-to-voxel splatting scheme.

\subsection{Cohesive Temporal Modeling}
\subsubsection{View-centric Temporal Modeling}
The view-centric path is designed to extract fine-grained temporal cues from historical observation sequences. By leveraging geometric 
constraints and ego-motion compensation, it adaptively samples spatio-temporal features from the image feature queue $\mathcal{I}$ and aggregates 
them into a unified 4D representation. 

Specifically, for each query $\mathbf{q}$ and its corresponding 3D reference point $\mathbf{p}_{ref}=(x,y,z)^T$, we first utilize a linear layer 
to predict a set of sampling offsets:
\begin{equation}
    \{(\triangle x^{i}, \triangle y^{i}, \triangle z^{i})\}_{i=1}^{N_s} = \mathrm{Linear}(\mathbf{q}).
\end{equation}                                         
These offsets are subsequently added to the reference point to generate a set of 3D sampling points:
\begin{equation}
    \mathcal{P}_s=\{\mathbf{p}^i=(x + \triangle x^{i}, y + \triangle y^{i}, z + \triangle z^{i})^T \}_{i=1}^{N_s}.
\end{equation}                                                
This sampling strategy allows queries to adaptively adjust the receptive field to capture more discriminative features. 

To sample features from any time $\tau \in \{t, t-1, \cdots, t-T+1 \}$ within the memory queue $\mathcal{I}$, we project each sampling point onto the hit camera views based on ego-motion 
compensation and camera parameters. The sampled feature $f^{\tau,i}$ for the $i$-th point at time $\tau$ is extracted via bilinear interpolation 
as follows:
\begin{equation}
    f^{\tau,i} = \frac{1}{|V_{hit}|} \sum\limits_{v\in V_{hit}}\sum\limits_{l=1}^{L} w^{i,l} \cdot \mathcal{BS}(I^{\tau,v,l}, \pi_{v}(T_{t \rightarrow{\tau}}\mathbf{p}^i)),
\end{equation}             
where $V_{hit}$ is the set of  hit camera views, $T_{t \rightarrow{\tau}} \in \mathbb{R}^{4\times4}$ denotes the ego-motion transformation 
matrix that transforms the 3D coordinates from the coordinate frame at current time $t$ to the target frame at time $\tau$.
$\pi_{v}(\cdot)$ denotes the projection function onto the $v$-th image plane, and $\mathcal{BS}(\cdot)$ denotes the bilinear sampling function. 
The term $w^{i,l}$ represents the weight predicted by the query for the $i$-th sampling point on the $l$-th feature scale.

To aggregate the spatio-temporal evidence from different timestamps and sampling points, we organize the extracted features into 
$f_{st} \in \mathbb{R}^{M \times C}$, where $M=T \times N_s$ represents the total number of sampled features across $T$ frames. 
Then, adaptive mixing~\cite{liu2023sparsebev, wang2024opus} is employed to transform $f_{st}$ along the channel and spatial dimensions 
under the guidance of the query. The resulting features are then aggregated into the query feature to produce an updated representation.

\subsubsection{Object-centric Temporal Modeling}
Sparse queries compactly encode the geometric and semantic information of the scene. Given the inherent temporal continuity of the 
evolution of driving scenarios, sparse query representations in adjacent frames exhibit strong correlation. Based on this observation, 
object-centric temporal modeling propagates these queries across frames, enabling efficient exploitation of informative historical 
spatial and semantic priors.


Specifically, we maintain a memory queue of size $N_p$ to store selected queries from the previous frame. Since these selected queries
are expected to predominantly cover foreground regions, an effective selection strategy is crucial. 
In our formulation, the occupancy field associated with each query is determined by its predicted superquadric parameters. 
In particular, the scale vector $\mathbf{s} = (s_x, s_y, s_z)^T$ controls the spatial extent of the field and thus reflects 
the geometric contribution of the query to the foreground. A larger scale typically indicates a more substantial contribution. 
Thus, we define the foreground score $\zeta$ of each query as the maximum scale among its $K$ predicted superquadrics: 
\begin{equation}
    \zeta = \max_{k \in \{1,\dots,K\}} \left( \max(s_x^{k}, s_y^{k}, s_z^{k}) \right).
\end{equation}


At each frame, the top-$N_p$ queries with the highest foreground scores and their reference points $P_{p}$ are pushed into a memory queue.
After temporal alignment, they are propagated to the next frame.
To alleviate potential spatial redundancy between propagated queries and newly initialized ones, we discard initialized queries whose reference 
points lie within a distance threshold $\delta$~(e.g., $\delta=1.0\text{ m}$) of any propagated reference point.
Then, the query set is complemented by stochastically sampling $N_q-N_p$ candidates from the residual initialization queries to 
maintain a constant budget $N_q$.
This ensures that the hybrid queries provide both stable temporal priors for persistent objects and sufficient exploratory coverage for newborn objects.

\subsection{Multi-Superquadric Decoding Strategy}

Existing superquadric-based methods~\cite{zuo2025quadricformer,hayes2025superquadricocc} typically decode each query into a single superquadric primitive. 
However, a single superquadric struggles to accurately represent irregular local geometry in real-world driving scenes, which forces the model to rely on 
high-density queries to achieve adequate geometric fidelity. 
To amplify the geometric expressiveness of the model without compromising the sparsity of the query representation, we propose the multi-superquadric decoding strategy. 
Under this strategy, each query $\mathbf{q}$ predicts a local superquadric cluster $\mathcal{S}_{local}$ composed of $K$ superquadrics:
\begin{equation}
\scalebox{0.95}{$
    \mathcal{S}_{local} = \{\mathbf{S}^{k}\}_{k=1}^K=\{(\mathbf{p}_{ref} +\mathbf{o}^{k}, \mathbf{r}^{k}, \mathbf{s}^{k}, \bm{\epsilon}^{k}, \sigma^{k}, \mathbf{c})\bigr\}_{k=1}^{K},
$}
\end{equation}
where $\mathbf{o}^k$ represents the predicted offset of the $k$-th superquadric center relative to the reference point $\mathbf{p}_{ref}$. 
All geometric parameters together with the opacity are generated by a regression head $\phi_{\text{reg}}$:
\begin{equation}
    \{(\mathbf{o}^{k}, \mathbf{r}^{k}, \mathbf{s}^{k}, \bm{\epsilon}^{k}, a^{k})\}_{k=1}^{K} = \phi_{\text{reg}}(\mathbf{q}).
\end{equation}
The semantic logits ${\mathbf{c}}$ are shared across the cluster and predicted by a classification head $\phi_{\text{cls}}$:
\begin{equation}
    {\mathbf{c}}=\phi_{\text{cls}}(\mathbf{q}).
\end{equation}

Furthermore, we employ a coarse-to-fine prediction mechanism to guide the decoder in smoothly transitioning from coarse structural approximation to fine-grained geometric refinement. 
Specifically, as the decoding depth increases, the number of superquadrics $K$ predicted by each query is progressively increased.

\subsection{Efficient Superquadric-to-Voxel Splatting}

The superquadric-to-voxel splatting aims to transform the predicted superquadrics into a volumetric semantic occupancy representation via a 
local aggregation operator. Considering the locality of the superquadric distribution, QuadricFormer~\cite{zuo2025quadricformer} first estimates the splatting radius based on the 
superquadric scale $\mathbf{s}$, and then determines the contributing superquadrics for each voxel through a matching and sorting procedure. 
During the splatting phase, a dedicated thread is assigned to each voxel to compute its occupancy probability and semantic logits.

However, this voxel-level binning strategy suffers from two major bottlenecks:
1)~The massive number of matching pairs formed between superquadrics and voxels significantly escalates the computational complexity of the 
sorting process.
2)~Neighboring voxels typically interact with highly similar sets of superquadrics. However, their corresponding threads necessitate redundant 
global memory transactions, leading to substantial memory bandwidth overhead.

To address these issues, inspired by the tile-based rasterization in gaussian splatting~\cite{kerbl20233d}, we partition the entire 3D space into cubic 
tiles of size $4\times4\times4$ voxels and perform binning operation at the tile level. 
This fundamentally reduces the population of matching pairs, thereby alleviating the sorting complexity. 
During the splatting stage, each tile is processed by a CUDA thread block, and each thread within the block still handles one voxel. 
Crucially, all superquadrics associated with a tile are coalesced and loaded into shared memory for collective reuse by all threads in the block. 
This mechanism drastically suppresses global memory traffic and significantly enhances overall computational throughput.
\section{Experiments}

\subsection{Datasets and Evaluation Metric}
nuScenes~\cite{caesar2020nuscenes} is a large-scale autonomous driving dataset that provides multi-sensor data collected from six cameras, one LiDAR, and five radars. 
It consists of 1,000 driving sequences, which are divided into 700 for training, 150 for validation, and 150 for testing. 
Each sequence lasts approximately 20 seconds, with a keyframe sampled every 0.5 seconds for annotation. 
The Occ3D~\cite{caesar2020nuscenes} and SurroundOcc~\cite{wei2023surroundocc} benchmarks provide dense semantic occupancy annotations for the 
nuScenes dataset. The occupancy annotations in Occ3D span [-40m, 40m] along the X and Y axes, [-1m, 5.4m] along the Z axis, with a resolution 
of $200 \times 200 \times 16$. It contains 17 semantic classes and one empty class. 
In comparison, SurroundOcc extends the spatial range to [-50m, 50m] along the X and Y axes and [-5m, 3m] along the Z axis, 
while maintaining the same $200 \times 200 \times 16$ resolution. Each voxel is labeled as one of 18 classes, including 16 semantic classes, 
one empty class, and one unknown class.

We adopt two evaluation metrics for SurroundOcc: the Intersection-over-Union~(IoU) of occupied voxels ignoring semantic classes, and the mean 
IoU~(mIoU) across all semantic classes. They are defined as:
\begin{equation}
    \mathrm{IoU}= \frac{TP_{\neq{c_0}}}{TP_{\neq{c_0}} + FP_{\neq{c_0}} + FN_{\neq{c_0}}},
\end{equation}
\begin{equation}
    \mathrm{mIoU}=\frac{1}{\mathcal{C'}} \sum \limits_{i \in \mathcal{C'}}\frac{TP_{i}}{TP_{i} + FP_{i} + FN_{i}},
\end{equation}
where ${\mathcal{C'}}$, $c_0$, $TP$, $FP$, $FN$ denote the nonempty classes, the empty class, the number of true positive, false positive and 
false negative predictions, respectively.
For Occ3D, we report the mIoU and the Ray-based IoU~(RayIoU)~\cite{liu2024sparseocc}. RayIoU is computed at three distance thresholds of 1, 2, and 4 meters, denoted 
as $\text{RayIoU}_{1m}$, $\text{RayIoU}_{2m}$, and $\text{RayIoU}_{4m}$. 
The final RayIoU score is obtained by averaging these three values.

\subsection{Implementation Details}
Following previous works~\cite{liu2024sparseocc, wang2024opus}, the input images are resized to $704\times256$. We adopt a ResNet-50~\cite{he2016deep} backbone with a 
FPN~\cite{lin2017feature} neck as the image encoder, where the backbone is initialized by nuImages~\cite{caesar2020nuscenes} pre-training.
To explore the trade-off between efficiency and performance, we scale our model into four variants, namely SuperOcc-T, SuperOcc-S, SuperOcc-M, and SuperOcc-L. 
These variants employ 0.6K, 1.2K, 2.4K, and 3.6K queries, respectively.  The specific configurations for each variant are detailed in Tab.~\ref{tab:setting}.

During training, we employ the AdamW~\cite{loshchilov2017decoupled} optimizer with a weight decay of $1 \times 10^{-2}$ and a batch size of 8. 
The learning rate is initialized at $2 \times 10^{-4}$ and decayed with a cosine annealing schedule. 
For comparison with SOTA methods, the models are trained for 48 epochs on Occ3D~\cite{tian2023occ3d} and 24 epochs on SurroundOcc~\cite{wei2023surroundocc}, respectively. 
For ablation studies, all experiments are conducted using the SuperOcc-T variant and trained for 24 epoch on Occ3D. 
FPS is evaluated on a single NVIDIA RTX 4090 GPU using the PyTorch FP32 backend with a batch size of 1.

\begin{table}[t]
    \small
    \caption{Configurations for different variants}
    \label{tab:setting}
    \centering
    \resizebox{0.5\textwidth}{!}{
    \begin{tabular}{l|ccc|cccccc}
    \toprule
    \multirow{2}{*}{Model} & \multirow{2}{*}{$N_q$}  & \multirow{2}{*}{$N_s$} & \multirow{2}{*}{$N_p$}  & \multicolumn{6}{c}{superquadric number}   \\
                           &                       &            &          & K1 & K2 & K3 & K4 & K5 & K6  \\ \midrule
    SuperOcc-T             & 600                   & 4          &   500    & 2  & 2  & 4  & 4  & 8  & 8   \\
    SuperOcc-S             & 1200                  & 2          &   1000   & 2  & 2  & 4  & 4  & 8  & 8   \\
    SuperOcc-M             & 2400                  & 2          &   2000   & 1  & 1  & 2  & 2  & 4  & 4   \\
    SuperOcc-L             & 3600                  & 2          &   3000   & 1  & 1  & 2  & 2  & 4  & 4   \\ 
    \bottomrule
    \end{tabular}}
\end{table}
\begin{table*}[h]
    \small
    \caption{
        \textbf{Occupancy prediction performance on Occ3D\cite{tian2023occ3d} benchmark}
        "8f" and "16f" denote models fusing temporal information from 8 or 16 frames, respectively.
        FPS results are measured on a single RTX 4090 GPU
    }
    \label{tab:occ3d}
    \centering
    \resizebox{\textwidth}{!}{
        \begin{tabular}{l|ccc|c|c|ccc|>{\columncolor{lightgray!20}}c|>{\columncolor{lightgray!20}}c}
        \toprule[0.9pt]
        Methods                                     & Backbone & Image Size      & Epoch  & Vis.Mask   & mIoU & $\text{RayIoU}_\text{1m}$ & $\text{RayIoU}_\text{2m}$ & $\text{RayIoU}_\text{4m}$ & RayIoU & FPS \\ \midrule
        RenderOcc~\cite{pan2024renderocc}           & Swin-B   & $1408\times512$ & 12     & \checkmark & 24.5 & 13.4      & 19.6      & 25.5      & 19.5    & -    \\
        BEVFormer~\cite{li2022bevformer}            & R101     & $1600\times900$ & 24     & \checkmark & 39.3 & 26.1      & 32.9      & 38.0      & 32.4    & 4.4  \\
        BEVDer-Occ~(2f)\cite{huang2021bevdet}       & R50      & $704\times256$  & 90     & \checkmark & 36.1 & 23.6      & 30.0      & 35.1      & 29.6    & 9.0  \\
        BEVDer-Occ~(8f)~\cite{huang2021bevdet}      & R50      & $704\times384$  & 90     & \checkmark & 39.3 & 26.6      & 33.1      & 38.2      & 32.6    & 5.6  \\
        SparseOcc~(8f)~\cite{liu2024sparseocc}      & R50      & $704\times256$  & 24     & -          & 30.9 & 28.0      & 34.7      & 39.4      & 34.0    & 20.9 \\
        SparseOcc~(16f)~\cite{liu2024sparseocc}     & R50      & $704\times256$  & 24     & -          & 30.6 & 29.1      & 35.8      & 40.3      & 35.1    & 15.9 \\
        P-FlashOcc-Tiny~(1f)~\cite{yu2024panoptic}  & R50      & $704\times256$  & 24     & -          & 29.1 & 29.1      & 35.7      & 39.7      & 34.8    & 41.9 \\
        P-FlashOcc~(1f)~\cite{yu2024panoptic}       & R50      & $704\times256$  & 24     & -          & 29.4 & 29.4      & 36.0      & 40.1      & 35.2    & 33.8 \\
        P-FlashOcc~(2f)~\cite{yu2024panoptic}       & R50      & $704\times256$  & 24     & -          & 30.3 & 31.2      & 37.6      & 41.5      & 36.8    & 29.8 \\
        P-FlashOcc~(8f)~\cite{yu2024panoptic}       & R50      & $704\times256$  & 24     & -          & 31.6 & 32.8      & 39.3      & 43.4      & 38.5    & 26.8 \\
        GSD-Occ~(16f)~\cite{he2025gsdocc}           & R50      & $704\times256$  & 24     & -          & 31.8 & -         & -         & -         & 38.9    & -    \\ 
        FB-Occ~(16f)~\cite{li2023fbocc}             & R50      & $704\times256$  & 24     & -          & 31.1 & 33.0      & 40.0      & 44.0      & 39.0    & 10.3 \\
        OPUS-T~(8f)~\cite{wang2024opus}             & R50      & $704\times256$  & 100    & -          & 33.2 & 31.7      & 39.2      & 44.3      & 38.4    & 25.1 \\
        OPUS-S~(8f)~\cite{wang2024opus}             & R50      & $704\times256$  & 100    & -          & 34.2 & 32.6      & 39.9      & 44.7      & 39.1    & 24.5 \\
        OPUS-M~(8f)~\cite{wang2024opus}             & R50      & $704\times256$  & 100    & -          & 35.6 & 33.7      & 41.1      & 46.0      & 40.3    & 17.8 \\
        OPUS-L~(8f)~\cite{wang2024opus}             & R50      & $704\times256$  & 100    & -          & 36.2 & 34.7      & 42.1      & 46.7      & 41.2    & 8.6  \\
        STCOcc~(8f)~\cite{liao2025stcocc}           & R50      & $704\times256$  & 36     & -          & -    & 36.2      & 42.7      & 46.4      & 41.7    & -    \\
        ALOcc-2D-mini~(16f)~\cite{chen2025alocc}    & R50      & $704\times256$  & 54     & -          & 33.4 & 32.9      & 40.1      & 44.8      & 39.3    & 30.5 \\
        ALOcc-2D~(16f)~\cite{chen2025alocc}         & R50      & $704\times256$  & 54     & -          & 37.4 & 37.1      & 43.8      & 48.2      & 43.0    & 8.2 \\      
        ALOcc-3D~(16f)~\cite{chen2025alocc}         & R50      & $704\times256$  & 54     & -          & 38.0 & 37.8      & 44.7      & 48.8      & 43.7    & 6.0 \\
        \midrule
        SuperOcc-T~(8f)       & R50      & $704\times256$  & 48 & -      & 36.1 & 36.6      & 43.4    & 47.6  & 42.5 & 30.3 \\
        SuperOcc-S~(8f)       & R50      & $704\times256$  & 48 & -      & 36.9 & 37.2      & 43.7    & 48.0  & 43.0 & 28.2 \\
        SuperOcc-M~(8f)       & R50      & $704\times256$  & 48 & -      & 37.4 & 37.8      & 44.4    & 48.6  & 43.6 & 18.8 \\
        SuperOcc-L~(8f)       & R50      & $704\times256$  & 48 & -      & 38.1 & 38.3      & 44.8    & 48.9  & 44.0 & 12.7 \\
        \bottomrule
        \end{tabular}
    }
\end{table*}

\subsection{Main Results}
\begin{table*}[t] %
    \caption{\textbf{3D semantic occupancy prediction results on SurroundOcc~\cite{wei2023surroundocc} benchmark}.
    * means supervised by dense occupancy annotations as opposed to original LiDAR segmentation labels.
    FPS results are measured on a single RTX 4090 GPU}
    \label{tab:surroundocc}
    \small
    \setlength{\tabcolsep}{0.005\linewidth}  
    \vspace{-2mm}  
    \renewcommand\arraystretch{1.05}
    \centering
    \resizebox{\textwidth}{!}{
    \begin{tabular}{l|c c | c c c c c c c c c c c c c c c c|c}
        \toprule
        Method
        & IoU
        & mIoU
        & \rotatebox{90}{\textcolor{nbarrier}{$\blacksquare$} barrier}
        & \rotatebox{90}{\textcolor{nbicycle}{$\blacksquare$} bicycle}
        & \rotatebox{90}{\textcolor{nbus}{$\blacksquare$} bus}
        & \rotatebox{90}{\textcolor{ncar}{$\blacksquare$} car}
        & \rotatebox{90}{\textcolor{nconstruct}{$\blacksquare$} const. veh.}
        & \rotatebox{90}{\textcolor{nmotor}{$\blacksquare$} motorcycle}
        & \rotatebox{90}{\textcolor{npedestrian}{$\blacksquare$} pedestrian}
        & \rotatebox{90}{\textcolor{ntraffic}{$\blacksquare$} traffic cone}
        & \rotatebox{90}{\textcolor{ntrailer}{$\blacksquare$} trailer}
        & \rotatebox{90}{\textcolor{ntruck}{$\blacksquare$} truck}
        & \rotatebox{90}{\textcolor{ndriveable}{$\blacksquare$} drive. suf.}
        & \rotatebox{90}{\textcolor{nother}{$\blacksquare$} other flat}
        & \rotatebox{90}{\textcolor{nsidewalk}{$\blacksquare$} sidewalk}
        & \rotatebox{90}{\textcolor{nterrain}{$\blacksquare$} terrain}
        & \rotatebox{90}{\textcolor{nmanmade}{$\blacksquare$} manmade}
        & \rotatebox{90}{\textcolor{nvegetation}{$\blacksquare$} vegetation}
        & FPS
        \\
        \midrule
        MonoScene~\cite{cao2022monoscene}       & 23.96 & 7.31 & 4.03 &	0.35& 8.00& 8.04&	2.90& 0.28& 1.16&	0.67&	4.01& 4.35&	27.72&	5.20& 15.13&	11.29&	9.03&	14.86 & -\\
        
        Atlas~\cite{murez2020atlas}             & 28.66 & 15.00 & 10.64&	5.68&	19.66& 24.94& 8.90&	8.84&	6.47& 3.28&	10.42&	16.21&	34.86&	15.46&	21.89&	20.95&	11.21&	20.54 & -\\
        
        BEVFormer~\cite{li2022bevformer}        & 30.50 & 16.75 & 14.22 &	6.58 & 23.46 & 28.28& 8.66 &10.77& 6.64& 4.05& 11.20&	17.78 & 37.28 & 18.00 & 22.88 & 22.17 & {13.80} & 22.21 & 3.3 \\
        
        TPVFormer~\cite{huang2023tri}           & 11.51 & 11.66 & 16.14&	7.17& 22.63	& 17.13 & 8.83 & 11.39 & 10.46 & 8.23&	9.43 & 17.02 & 8.07 & 13.64 & 13.85 & 10.34 & 4.90 & 7.37 & 2.9 \\
        
        TPVFormer*~\cite{huang2023tri}          & {30.86} & 17.10 & 15.96&	 5.31& 23.86	& 27.32 & 9.79 & 8.74 & 7.09 & 5.20& 10.97 & 19.22 & {38.87} & {21.25} & {24.26} & {23.15} & 11.73 & 20.81 & 2.9 \\

        OccFormer~\cite{zhang2023occformer}     & {31.39} & {19.03} & {18.65} & {10.41} & {23.92} & {30.29} & {10.31} & {14.19} & {13.59} & {10.13} & {12.49} & {20.77} & {38.78} & 19.79 & 24.19 & 22.21 & {13.48} & {21.35} & -\\
         
        SurroundOcc~\cite{wei2023surroundocc}   & {31.49} & {20.30}  & {20.59} & {11.68} & 28.06 & 30.86 & {10.70} & {15.14} & 14.09 & 12.06 & 14.38 & 22.26 & 37.29 & {23.70} & {24.49} & {22.77} & {14.89} & {21.86} & 3.3 \\

        GaussianFormer~\cite{huang2024gaussianformer}   & 29.83 & {19.10} & {19.52} & {11.26} & {26.11} & {29.78} & {10.47} & {13.83} & {12.58} & {8.67} & {12.74} & {21.57} & {39.63} & {23.28} & {24.46} & {22.99} & 9.59 & 19.12 & 2.7\\

        GaussianFormer-2~\cite{huang2025gaussianformer} & 30.56 & {20.02} & 20.15 & {12.99} & {27.61} & {30.23} & {11.19} & {15.31} & {12.64} & {9.63} & {13.31} & {22.26} & {39.68} & {23.47} & {25.62} & {23.20} & 12.25 & 20.73 & 2.8 \\

        QuadricFormer~\cite{zuo2025quadricformer}       & {31.22} & {20.12} & 19.58 & {13.11} & 27.27 & 29.64 & {11.25} & {16.26} & 12.65 & 9.15 & 12.51 & 21.24 & {40.20} & {24.34} & {25.69} & {24.24} & 12.95 & 21.86 & 6.2\\
        
        GaussianWorld~\cite{zuo2025gaussianworld}       & 33.40 & 22.13 & 21.38 & 14.12 & 27.71 & 31.84 & 13.66 & 17.43 & 13.66 & 11.46 & 15.09 & 23.94 & 42.98 & 24.86 & 28.84 & 26.74 & 15.69 & 24.74 & 4.4 \\

        \midrule
        SuperOcc-T & {34.91} & {22.48} & 21.35 & 11.90 & 28.35 & 32.12 & 15.46 & 15.84 & 13.20 & 11.83 & 11.87 & 23.71 & 47.30 & 29.19 & 30.29 & 28.22 & 13.23 & 25.75 & 30.3 \\
        SuperOcc-S & {35.63} & {23.12} & 21.96 & 12.14 & 28.46 & 32.86 & 16.98 & 16.62 & 13.60 & 13.06 & 12.40 & 24.52 & 47.16 & 29.07 & 30.81 & 28.46 & 14.90 & 26.87 & 28.2 \\
        SuperOcc-M & {36.99} & {23.95} & 23.73 & 12.07 & 28.46 & 33.13 & 17.42 & 17.68 & 14.62 & 14.20 & 12.87 & 25.27 & 47.84 & 29.33 & 31.58 & 29.15 & 17.28 & 28.55 & 18.8 \\
        SuperOcc-L & {38.13} & {24.55} & 23.92 & 12.43 & 29.77 & 33.62 & 17.36 & 17.74 & 14.95 & 14.79 & 13.42 & 26.25 & 48.22 & 29.22 & 32.16 & 30.20 & 18.67 & 30.04 & 12.7 \\
        \bottomrule
    \end{tabular}}
    \label{tab: nuscenes results}
    \vspace{-4mm}
\end{table*}
\subsubsection{Comparison on Occ3D}
We compare SuperOcc with recent occupancy prediction methods on the Occ3D~\cite{tian2023occ3d} benchmark. As reported in Tab.~\ref{tab:occ3d}, SuperOcc establishes new SOTA performance while achieving a 
superior trade-off between prediction accuracy and inference efficiency.

Compared with the representative sparse predictor OPUS~\cite{wang2024opus}, SuperOcc-T achieves 42.5\% RayIoU and 30.3 FPS, surpassing OPUS-L in prediction accuracy and outperforming OPUS-T in 
inference speed. 
By scaling up the model, SuperOcc-S/M/L achieve further performance improvements. In particular, SuperOcc-L reaches 38.1\% mIoU and 44.0\% RayIoU, outperforming the previous SOTA ALOcc-3D~\cite{chen2025alocc} 
by 0.3\% RayIoU while maintaining nearly twice the inference throughput of ALOcc-3D~(12.7 vs. 6.0 FPS).
Moreover, these results demonstrate the strong scalability of SuperOcc: by simply adjusting parameters such as the number of queries, the model achieves a flexible and controllable trade-off 
between accuracy and efficiency. This enables SuperOcc to meet diverse real-world requirements for both real-time performance and high prediction accuracy.

\subsubsection{Comparison on SurroundOcc}
In Tab.~\ref{tab:surroundocc}, we present a comprehensive comparison with other SOTA 3D semantic occupancy prediction methods on the SurroundOcc benchmark. 
Our SuperOcc family consistently demonstrates a commanding performance lead over existing approaches.

Specifically, compared with the advanced temporal-aware predictor GaussianWorld~\cite{zuo2025gaussianworld}, SuperOcc achieves higher IoU and mIoU even under a weaker setting~(ResNet-50~\cite{he2016deep} vs. ResNet101-DCN~\cite{he2016deep}, and 
input resolution $256 \times 704$ vs. $900 \times 1600$). 
Furthermore, compared with the recent superquadric-based predictor QuadricFormer~\cite{zuo2025quadricformer}, our method also exhibits significant performance margins. 
In particular, SuperOcc-T outperforms QuadricFormer by 3.69\% IoU and 2.36\% mIoU, while providing substantially higher inference throughput. 
These results collectively establish SuperOcc as a more effective and efficient superquadric-based occupancy prediction framework.

\subsection{Ablation Studies}

\begin{table}[t]
    \centering
    \caption{
        Ablation of different components in SuperOcc.
        MDS:~Multi-Superquadric Decoding;
        CTM:~Cohesive Temporal Modeling
        }
    \label{tab:abl_component}
    \resizebox{0.9\columnwidth}{!}{
        \begin{tabular}{cc|cc|ccc|c}
        \toprule
        MSD & CTM & mIoU & RayIoU & \multicolumn{3}{c|}{RayIoU\textsubscript{1m, 2m, 4m}} & FPS \\
        \midrule
                    &             & 27.9 & 33.3 & 26.9 & 34.0 & 39.0 & \textbf{33.0} \\
        \checkmark  &             & 29.1 & 34.9 & 28.6 & 35.6 & 40.5 & 31.7 \\
                    & \checkmark  & 33.8 & 39.5 & 33.4 & 40.4 & 44.8 & 31.2 \\
        \checkmark  & \checkmark  & \textbf{35.4} & \textbf{41.2} & \textbf{35.0} & \textbf{41.9} & \textbf{46.5} & 30.3 \\
        \bottomrule
        \end{tabular}
    }

\end{table}

\subsubsection{Component analysis}
We conduct a comprehensive ablation study in Tab.~\ref{tab:abl_component} to quantify the individual contribution of each proposed component 
in the SuperOcc framework. 

The baseline relies solely on single-frame image inputs and single-superquadric decoding.
Starting from the baseline, incorporating the proposed multi-superquadric decoding~(MSD) strategy yields an improvement of 1.2\% mIoU and 1.6\% RayIoU, with only 
a marginal increase in inference latency. This indicates that MSD effectively enhances geometric expressiveness while completely preserving 
the query sparsity. When equipping the baseline with the cohesive temporal modeling~(CTM) mechanism, we observe a substantial performance gain of 5.9\% mIoU and 6.2\% RayIoU. 
This remarkable performance leap underscores the effectiveness and necessity of our CTM mechanism. 
By combining all components, SuperOcc achieves a significant overall improvement of 7.5\% mIoU and 7.9\% RayIoU over the baseline. 
Although this configuration incurs a moderate reduction in FPS, it maintains a real-time inference speed of 30.3 FPS, offering a 
favorable trade-off between accuracy and efficiency.

\begin{table}[t]
    \centering
	\caption{Ablation of Multi-Superquadric Decoding}
    \label{tab:msd}
    \resizebox{0.95\columnwidth}{!}{
        \begin{tabular}{c|cc|ccc}
        \toprule
        K1-K6 & mIoU & RayIoU & \multicolumn{3}{c}{RayIoU\textsubscript{1m, 2m, 4m}} \\
        \midrule
        $[1, 1, 1, 1, 1, 1]$        & 27.9  & 33.3 & 26.9 & 34.0 & 39.0 \\
        $[2, 2, 2, 2, 2, 2]$        & 28.4  & 33.9 & 27.4 & 34.6 & 39.6 \\
        $[4, 4, 4, 4, 4, 4]$        & 28.8  & 34.6 & 28.4 & 35.3 & 40.2 \\
        $[8, 8, 8, 8, 8, 8]$        & 28.8  & 34.9 & \textbf{28.7} & 35.6 & 40.4 \\
        $[16, 16, 16, 16, 16, 16]$  & 28.7  & 34.5 & 28.2 & 35.1 & 40.0 \\
        \midrule
        $[2, 2, 4, 4, 8, 8]$        & \textbf{29.1}  & \textbf{34.9} & 28.6 & \textbf{35.6} & \textbf{40.5} \\
        \bottomrule
        \end{tabular}
    }
\end{table}
\subsubsection{Effect of Multi-Superquadric Decoding}
The number of superquadrics predicted per query is a critical factor in our MSD strategy. As summarized in Tab.~\ref{tab:msd}, we 
conduct an ablation study to analyze its impact. 

The first five rows report the performance when varying the number of predicted superquadrics without the coarse-to-fine mechanism. 
We observe that occupancy prediction performance improves steadily with the number of superquadrics, and begins to saturate at $K=8$.
This demonstrates that while MSD effectively improves geometric expressiveness, an optimal number of primitives is essential to balance 
representational capacity and optimization stability, with $K=8$ providing the best trade-off in our settings.

In the final row, we incorporate the coarse-to-fine strategy, where the number of predicted superquadrics is progressively increased 
across decoding stages. This configuration further boosts performance, achieving the highest mIoU of 29.1\% and RayIoU of 34.9\%. 
This suggests that the coarse-to-fine strategy facilitates the learning process by allowing the network to 
progressively optimize geometric representations from coarse structures to fine details.

\begin{table}[t]
    \centering
	\caption{Ablation of Cohesive Temporal Modeling}
    \label{tab:jtm}
    \resizebox{0.95\columnwidth}{!}{
        \begin{tabular}{cc|cc|ccc}
        \toprule
        View-centric & Object-centric & mIoU & RayIoU & \multicolumn{3}{c}{RayIoU\textsubscript{1m, 2m, 4m}} \\
        \midrule
                    &             & 29.1  & 34.9  & 28.6  & 35.6  & 40.5 \\
        \checkmark  &             & 34.5  & 39.6  & 33.6  & 40.4  & 44.9 \\
                    & \checkmark  & 31.6  & 37.6  & 31.3  & 38.3  & 43.2 \\
        \checkmark  & \checkmark  & \textbf{35.4}  & \textbf{41.2}  & \textbf{35.0}  & \textbf{41.9}  & \textbf{46.5} \\
        \bottomrule
        \end{tabular}
    }
\end{table}
\subsubsection{Effect of Cohesive Temporal Modeling}
To explore the impact of different temporal modeling strategies, we conduct a comprehensive ablation study as shown in Tab.~\ref{tab:jtm}.  
Without temporal modeling, the baseline achieves limited performance, reflecting the inherent difficulty of accurately predicting 3D occupancy 
from single-frame inputs. It can be observed that incorporating view-centric temporal modeling significantly boosts mIoU and RayIoU by 5.4\% and 
4.7\%, respectively. This demonstrates that the rich spatio-temporal context provided by historical observations is crucial for inferring the current occupancy. 
Meanwhile, object-centric temporal modeling also yields noticeable gains~(+2.5\% mIoU and +2.7\% RayIoU), suggesting that the spatial and 
semantic priors propagated via historical queries can effectively guide the current decoding process. When view-centric and object-centric 
temporal modeling are jointly applied, the model achieves the best performance across all evaluation metrics. This result strongly validates the 
complementarity of the two modeling strategies, demonstrating that their integration is essential for achieving accurate 3D occupancy prediction.

\begin{table}[t]
    \centering
    \caption{Ablation study on different query selection strategies in object-centric temporal modeling}
    \label{tab:query_selection}
    \resizebox{0.9\columnwidth}{!}{
        \begin{tabular}{l|cc|ccc}
            \toprule
            Selection Strategy & mIoU & RayIoU & \multicolumn{3}{c}{RayIoU\textsubscript{1m, 2m, 4m}} \\
            \midrule
            No selection         & 25.3 & 31.1 & 24.9 & 31.7 & 36.6 \\
            Semantic logit-based & 29.5 & 35.4 & 29.0 & 36.1 & 41.1 \\
            Opacity-based        & 29.3 & 35.6 & 29.2 & 36.2 & 41.2 \\
            Scale-based (ours)   & \textbf{31.6} & \textbf{37.6} & \textbf{31.3} & \textbf{38.3} & \textbf{43.2} \\
            \bottomrule
        \end{tabular}
    }
\end{table}

\subsubsection{Effect of Scale-based Query Selection}
Table~\ref{tab:query_selection} investigates various query selection strategies for object-centric temporal modeling. 
Directly propagating all queries results in a noticeable performance drop, as the propagated queries tend to dominate the optimization process and 
impede the effective learning of initialization queries. 
While strategies based on semantic logits or opacity partially mitigate this, they achieve suboptimal performance because neither semantic confidence nor opacity reliably 
reflects a query's foreground contribution. 
In contrast, our scale-based selection achieves the best performance. This indicates that scale-based selection is more effective at identifying foreground queries, 
thereby providing more stable and efficient temporal priors for subsequent frames.

\begin{table}[t]
  \centering
  \caption{Ablation of the Efficient Superquadric-to-Voxel Splatting}
  \label{tab:ess}
  \resizebox{0.95\columnwidth}{!}{
    \begin{tabular}{l|c|c|c|c|c}
      \toprule
      \multirow{2}{*}{ESS} & \multicolumn{2}{c|}{Training Phase} & \multicolumn{3}{c}{Inference Phase} \\
      & Mem (GB) & Time (h) & Mem (GB) & Splatting Latency (ms) & FPS \\
      \midrule
                & 21.4 & 82 & 3.1 & 6.2 & 25.2 \\
      \checkmark & \textbf{17.3} \scriptsize{(-19\%)} & \textbf{20} \scriptsize{(-76\%)} & \textbf{2.8} \scriptsize{(-9\%)} & \textbf{1.3} \scriptsize{(-79\%)} & \textbf{30.3} \scriptsize{(+20\%)} \\
      \bottomrule
    \end{tabular}
  }
\end{table}
\subsubsection{Effect of Efficient Superquadric-to-Voxel Splatting}
To evaluate the impact of the proposed efficient superquadric-to-voxel splatting~(ESS), we conduct a comparative analysis as summarized in Tab.~\ref{tab:ess}. The 
experimental results demonstrate that ESS significantly reduces computational costs across both the training and inference 
stages.

In the training phase, the integration of ESS significantly boosts training efficiency. Notably, it achieves a 76\% 
reduction in wall-clock training time (from 82 h to 20 h) while simultaneously lowering the GPU memory footprint by 19\%~(from 21.4 GB to 17.3 GB).
This drastic acceleration demonstrates that our optimized splatting operator effectively alleviates the computational bottlenecks in the 
superquadric-to-voxel transformation during both forward and backward propagation.
In the inference phase, ESS slashes the splatting latency by 79\%, from 6.2 ms to 1.3 ms. Consequently, the overall system throughput 
is increased by 20\%, reaching 30.3 FPS. This reduction in computational overhead significantly enhances the deployment potential of our framework.

\subsection{Qualitative Results}
\begin{figure*}[t]
  \centering
  \includegraphics[width=\linewidth]{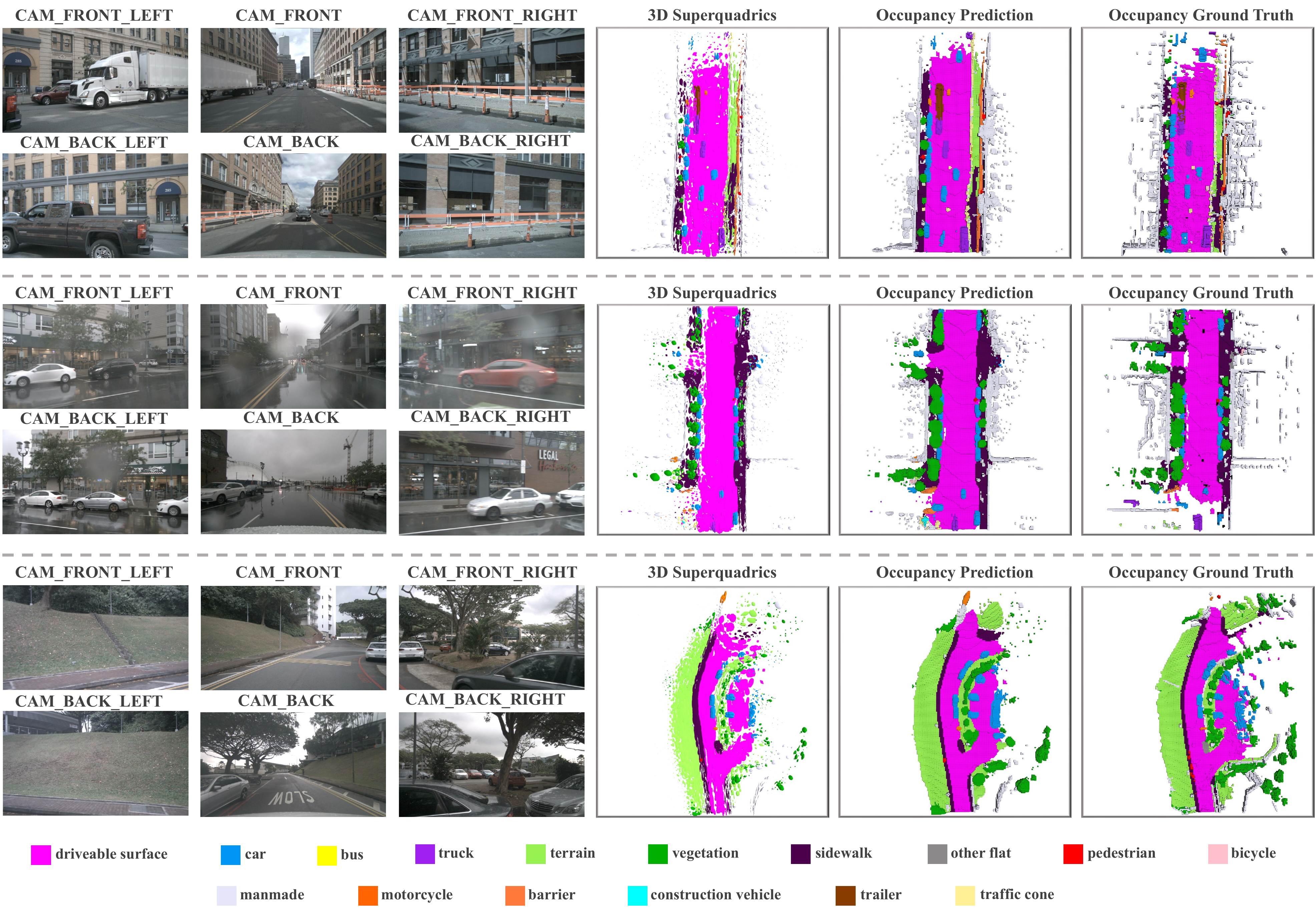}
  \caption{Qualitative results of SuperOcc on the SurroundOcc~\cite{wei2023surroundocc} benchmark. 
  For each scene, the predicted superquadrics, resulting occupancy prediction, and ground-truth occupancy are shown.
  Colors indicate semantic categories for both superquadrics and voxels.}
  \label{fig:vis}
\end{figure*}

Qualitative prediction results of SuperOcc on the SurroundOcc~\cite{wei2023surroundocc} benchmark are shown in Fig.~\ref{fig:vis}.
For each scene, we visualize the predicted superquadrics, the resulting occupancy prediction,
and the corresponding ground-truth occupancy. Each superquadric and voxel is color-coded
according to its semantic category.

The predicted superquadric primitives effectively model the underlying 3D scene geometry
in a compact manner. The occupancy predictions derived from these superquadrics exhibit
high fidelity to the ground truth, preserving clear structural boundaries between occupied
and free space. These results demonstrate that SuperOcc can accurately reconstruct
fine-grained scene geometries, highlighting its effectiveness in complex urban environments.

\subsection{Limitations}
Although our proposed method achieves significant progress, several limitations remain to be addressed.

First, SuperOcc is restricted to semantic occupancy prediction and has not yet been extended to panoptic occupancy 
prediction. Specifically, while it can accurately predict the semantic category of each occupied voxel, it lacks 
the ability to distinguish different individual instances within the same category. However, such instance-level 
discrimination is crucial for fine-grained scene understanding in autonomous driving.
Second, SuperOcc primarily relies on ego-motion compensation for temporal alignment, without explicitly modeling 
the motion of dynamic objects. As a result, historical features sampled for dynamic objects may exhibit spatio-temporal 
misalignment. Moreover, the temporal propagation of queries may introduce less reliable 
spatial priors, which can adversely affect the occupancy modeling of dynamic objects.

In future work, we plan to extend SuperOcc to panoptic occupancy prediction by incorporating instance-level representations 
into the superquadric-based framework. Additionally, we aim to integrate explicit object motion modeling to enhance the 
spatio-temporal consistency for dynamic objects.

\section{Conclusion}
In this paper, we propose SuperOcc, a novel superquadric-based framework for 3D occupancy prediction. By unifying view-centric and 
object-centric temporal paradigms, SuperOcc effectively exploits fine-grained spatio-temporal context together with historical priors, 
enabling accurate predictions in dynamic driving environments. Furthermore, SuperOcc introduces a multi-superquadric decoding strategy 
that decodes each query into a cluster of superquadrics. This design significantly bolsters the capacity to model complex scenes while 
preserving query sparsity. Additionally, we optimize the superquadric-to-voxel splatting operation through a CUDA-friendly implementation, 
dramatically reducing training overhead and splatting latency. Extensive evaluations on public benchmarks validate the effectiveness and 
efficiency of SuperOcc, demonstrating its clear advantages over existing occupancy prediction methods. We hope that SuperOcc can provide 
valuable insights into superquadric-based occupancy prediction for the community. 

\label{sec:conclusion}

\bibliographystyle{IEEEtran}
\bibliography{SuperOcc}

\end{document}